\documentclass[letterpaper, 10 pt, conference]{ieeeconf}  

\IEEEoverridecommandlockouts                              

\usepackage[utf8]{inputenc} 
\usepackage[T1]{fontenc}    
\usepackage{hyperref}       
\usepackage{url}            
\usepackage{booktabs}       
\usepackage{amsfonts}       
\usepackage{nicefrac}       
\usepackage{microtype}      
\usepackage{xcolor}         

\usepackage{graphicx}
\usepackage{comment}
\usepackage{amsmath,amssymb} 
\usepackage{color}
\usepackage{bm}
\usepackage{amsfonts}
\usepackage{array}
\usepackage{algorithm}
\usepackage{algorithmic}
\usepackage{cite}
\usepackage{booktabs}
\usepackage{float}
\usepackage{multirow}
\usepackage{mathrsfs}
\usepackage{amsfonts,amssymb} 
\usepackage{lipsum}
\usepackage{wrapfig}
\usepackage{makecell}
\usepackage{soul}

\usepackage{setspace}
\title{\LARGE \bf
Important Object Identification with Semi-Supervised Learning for Autonomous Driving
}

\author{Jiachen Li*, Haiming Gang*, Hengbo Ma, Masayoshi Tomizuka, \textit{Life Fellow, IEEE}, and Chiho Choi
\thanks{*The authors contributed equally to this work}
\thanks{Jiachen Li is with the Honda Research Institute, San Jose, CA 95134 USA, and also with the Department of Aeronautics \& Astronautics, Stanford University, CA 94305 USA (e-mail: jiachen\_li@stanford.edu).}
\thanks{Haiming Gang and Chiho Choi are with the Honda Research Institute, San Jose, CA 95134 USA (e-mail: haiminggang@gmail.com; cchoi@honda-ri.com).}
\thanks{Hengbo Ma and Masayoshi Tomizuka are with the Department of Mechanical Engineering, University of California, Berkeley, CA 94720 USA (e-mail: hengbo\_ma@berkeley.edu, tomizuka@me.berkeley.edu).}
}

\begin{document}

\maketitle
\thispagestyle{empty}
\pagestyle{empty}

\begin{abstract}
Accurate identification of important objects in the scene is a prerequisite for safe and high-quality decision making and motion planning of intelligent agents (e.g., autonomous vehicles) that navigate in complex and dynamic environments.
Most existing approaches attempt to employ attention mechanisms to learn importance weights associated with each object indirectly via various tasks (e.g., trajectory prediction), which do not enforce direct supervision on the importance estimation. 
In contrast, we tackle this task in an explicit way and formulate it as a binary classification ("important" or "unimportant") problem.
We propose a novel approach for important object identification in egocentric driving scenarios with relational reasoning on the objects in the scene.
Besides, since human annotations are limited and expensive to obtain, we present a semi-supervised learning pipeline to enable the model to learn from unlimited unlabeled data.
Moreover, we propose to leverage the auxiliary tasks of ego vehicle behavior prediction to further improve the accuracy of importance estimation.
The proposed approach is evaluated on a public egocentric driving dataset (H3D) collected in complex traffic scenarios. 
A detailed ablative study is conducted to demonstrate the effectiveness of each model component and the training strategy. Our approach also outperforms rule-based baselines by a large margin.
\end{abstract}

\section{INTRODUCTION}
The autonomous agents that navigate in complex, highly dynamic environments need to accurately identify the most important objects in the scene which are relevant to their decision making and motion planning.
In particular, autonomous vehicles should be able to figure out which \textit{dynamic objects} (e.g., vehicles, pedestrians, cyclists) and \textit{static objects} (e.g., traffic lights, stop signs) are critical to perceive the environment, detect potential risks, and determine the current action given their specific intentions/goals.
We define the important objects at the current frame as ``the objects that present semantics meanings that influence the ego behavior (e.g., traffic lights, stop signs), or the objects that may influence the ego behavior through interactions''.
Human drivers tend to focus on a subset of important surrounding objects when driving in a dense area.
Similarly, the limited onboard computational resources can be allocated more efficiently on the perception and reasoning of the identified important objects.
Moreover, this function can also enable the advanced driver-assistance systems to warn the drivers about risk objects in dangerous situations.

Existing related works can be mainly divided into three categories.
First, some works focus on predicting the driver's gaze by imitating human drivers \cite{martin2018dynamics,palazzi2018predicting}. The gaze information can be obtained by mounting a camera on the driver's head. However, most of these methods only provide pixel/region-level attention without indicating the importance of each object/instance. 
Driver gaze tends to be sequential and limited to a single region at a certain moment, while there may be multiple important objects out of the focused region simultaneously. Moreover, human drivers may not always pay attention to the truly important objects, thus degrades the reliability of the information.
Second, some works attempt to train an attention-based model with specific tasks such as trajectory forecasting \cite{vemula2018social,li2021rain,li2021spatio} and end-to-end driving \cite{aksoy2020see}, in which there is no explicit supervision on the learned attention. 
Third, some recent works attempt to identify important objects by providing explicit supervision on the object importance with human annotations to inject human knowledge, in which the models are trained by standard supervised learning \cite{gao2019goal,zhang2020interaction}. However, these approaches demand a large amount of labeled data. 
Our proposed method lies in this direction of research, but we propose to utilize semi-supervised learning techniques to reduce human efforts and enable the model to learn an unlimited amount of unlabeled data. Different from \cite{gao2019goal,zhang2020interaction} which only consider dynamic traffic participants, we also include traffic lights/signs in the driving scenes to enable semantic reasoning of the environment.

Learning object importance from human-labeled data can be naturally formulated as a binary classification problem, where each object is classified as important (1) or unimportant (0).
However, since the importance of a certain object is not totally independent from others (e.g., the importance of a certain object may decrease given the existence of another object), it is necessary to reason about the relation among entities before the final classification. 
Therefore, we leverage a graph representation and message passing operations to extract relational features.

Moreover, since the behavior of ego vehicle may be influenced by the important objects, it may in turn provide helpful cues for important object identification. 
For example, if the ego vehicle is waiting before a green light, there is likely at least one important object which prevents it from moving forward.
Therefore, we propose to use auxiliary tasks of ego behavior prediction to provide additional supervision signals for the importance classification.
Equipped with the auxiliary branches, our framework is able to identify important objects in the scene and infer the ego behavior simultaneously.

The main contributions of this paper are summarized as follows.
First, we propose a novel method for important object identification in the egocentric driving scenarios with relational reasoning on the objects in the scene.
Second, we propose to employ a modified semi-supervised learning algorithm with a ranking-based strategy for the pseudo-label generation to enable the model to learn from unlabeled datasets in addition to a human-labeled dataset. To the best of our knowledge, we are the first to apply a modified semi-supervised learning algorithm to important object/people identification for autonomous driving.
Third, we propose to leverage the auxiliary tasks of ego vehicle behavior prediction to provide additional supervision signals and further improve the accuracy of importance estimation.
Finally, we validate the proposed method on a public urban driving dataset with a detailed ablative study and comparison with baselines, which demonstrates the effectiveness of each model component and the training strategy.

\vspace{-0.cm}
\section{Related work}
\vspace{-0.cm}

\textbf{Important Object/People Identification}:
Identifying important individuals automatically from a set of entities has attracted increasing research efforts, which can be applied to a wide range of application domains.
While many visual saliency detection approaches have been proposed to solve various computer vision tasks such as visual question answering \cite{anderson2018bottom,yu2019deep}, scene understanding \cite{sun2020mining}, video summarization \cite{ji2020deep}, most of them only provide pixel-level importance estimation without being aware of individual object instances.
Some recent papers put more emphasis on the object/instance-level importance estimation and their applications.
In \cite{li2019learning}, Li et al. proposed a framework for automatically detecting important people in still images of social events.
In the autonomous driving domain, Li et al. \cite{li2020who} presented a model to identify risk objects for risk assessment via causal inference.
Gao et al. \cite{gao2019goal} proposed a method to estimate importance of traffic participants in egocentric driving videos, and Zhang et al. \cite{zhang2020interaction} further considered the interactions among the involved entities.
However, most of these models are trained by supervised learning which demand 
a large amount of labeled data.

\textbf{Semi-supervised Learning (SSL)}:
Learning from partially labeled data has emerged as an exciting research direction in deep learning, especially in classification problems \cite{zhou2019collaborative,jing2021videossl,hong2020learning}.
It enables the models to effectively learn from a labeled data together with an unlimited amount of unlabeled data, reducing the efforts of human annotation and enlarging learning resources \cite{ouali2020overview}.
The typical SSL methods can be broadly divided into the following categories: consistency regularization \cite{tarvainen2017mean}, proxy-label methods \cite{lee2013pseudo}, generative models \cite{kingma2014semi}, and graph-based methods \cite{iscen2019label}.
However, these SSL methods were primarily proposed for the standard classification tasks where the object instances are classified independently without considering their relations. 
In this work, we propose a modified strategy for pseudo-label generation and reason about the relations between objects.

\textbf{Relational Reasoning and Graph Neural Networks}:
In order to identify the important individuals in a given scene, the model should learn to recognize their relations.
Relational reasoning on a group of entities has a wide range of applications such as trajectory forecasting \cite{choi2020shared,li2020evolvegraph,ma2021continual,ma2021multi,zhou2022grouptron}, interaction detection \cite{zhou2019relation}, object detection \cite{hu2018relation}, dynamics modeling \cite{sanchez2020learning}, human-robot interaction \cite{chen2020relational}.
In recent years, graph neural networks \cite{battaglia2018relational} have attracted significantly increasing research efforts in various fields, which are suitable for tackling relation modeling and extraction \cite{alet2019neural,li2020evolvegraph}.
In this work, we employ a graph neural network to model the relations among objects in the driving scenes, which improves the performance of important object identification.

\begin{figure*}[!tbp]
	\centering
	\includegraphics[width=0.85\textwidth]{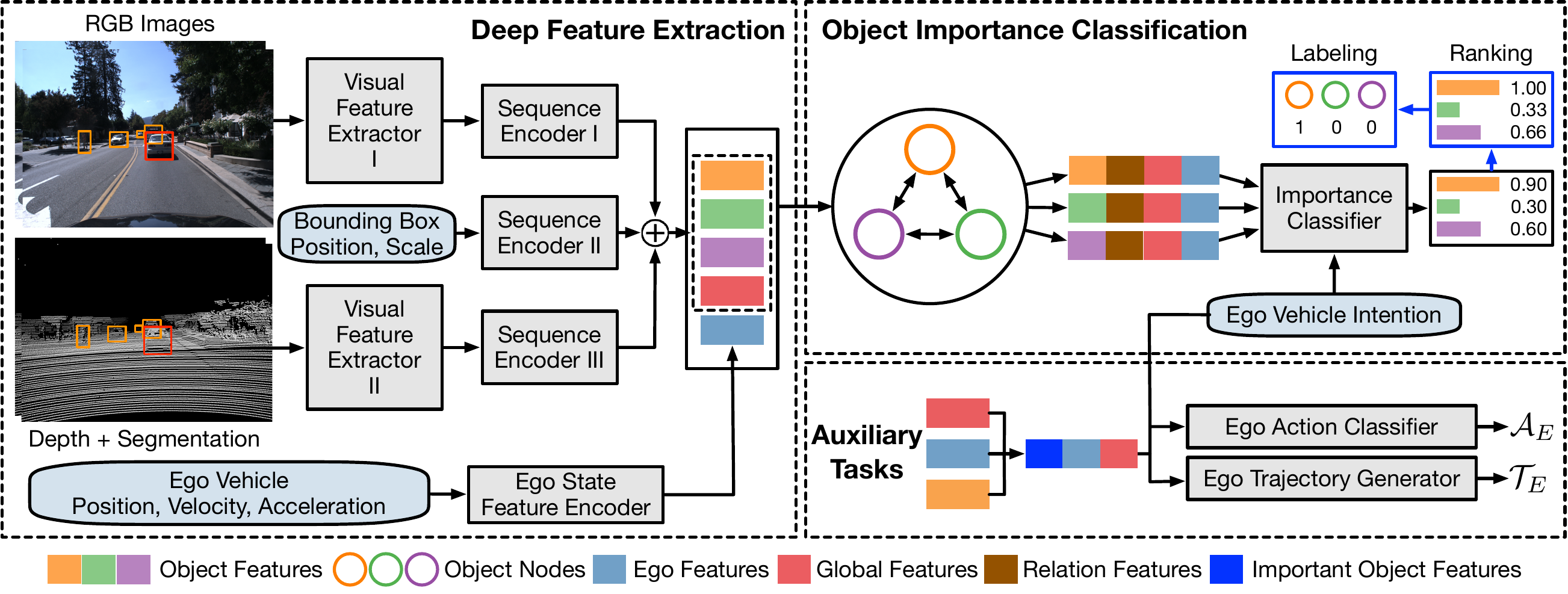}
	\vspace{-0.2cm}
	\caption{The framework of important object identification with a semi-supervised learning pipeline, which consists of three components: 1) deep feature extraction (DFE); 2) object importance classification (OIC); and 3) auxiliary tasks (AT). The procedures with blue arrows are only applied to unlabeled data samples for pseudo-label generation. In the input images, the red/orange boxes indicate the important/unimportant objects, respectively.}
	\vspace{-0.6cm}
	\label{fig:model}
\end{figure*}

\vspace{-0.cm}
\section{Problem formulation}
\vspace{-0.cm}

The important object identification is formulated as a binary classification problem with a semi-supervised learning pipeline.
Consider a labeled egocentric driving dataset that contains $|\mathcal{L}|$ labeled frontal image sequences $\mathcal{L} = \{\mathbf{I}^\mathcal{L}_{i,t}, i=1,...,|\mathcal{L}|, t=-T_h+1,...,0\}$, where for image sequence $\mathbf{I}^\mathcal{L}_{i,t}$ there are $N^{\mathcal{L}}_i$ detected objects $\{\mathbf{x}^\mathcal{L}_{j},j=1,...,N^{\mathcal{L}}_i\}$ at the current frame $t=0$ and the corresponding importance labels $y^\mathcal{L}_j$. 
We set $y^\mathcal{L}_j=1$ for ``important'' class and $y^\mathcal{L}_j=0$ for ``unimportant'' class.
We also have a set of unlabeled frontal image sequences $\mathcal{U} = \{\mathbf{I}^\mathcal{U}_{i,t}, i=1,...,|\mathcal{U}|, t=-T_h+1,...,0\}$, where for image sequence $\mathbf{I}^\mathcal{U}_{i,t}$ there are $N^{\mathcal{U}}_i$ detected objects $\{\mathbf{x}^\mathcal{U}_{j},j=1,...,N^{\mathcal{U}}_i\}$ at the current frame $t=0$ without importance labels.
In this paper, we aim to design an importance classification model $\hat{\mathbf{y}} = f_{\theta}(\mathbf{x}_j)$ where $\mathbf{x}_j \in \mathbf{I}_{i,0}$ to learn from the augmented training set $\mathcal{L}\cup \mathcal{U}$. 
In other words, we care about the objects existing at the current frame, and the model $f_{\theta}$ takes in the information of all the detected objects and predicts their importance with respect to the ego vehicle.
Besides, we have access to other onboard sensor measurements (e.g., point cloud, velocity, acceleration, yaw angle) and the current ego driving intention $I_E$ (i.e., going forward, turning left, turning right).

\section{Method}
An illustrative model diagram is shown in Fig. 1 to demonstrate the essential components and procedures of the important object identification approach. 
The model consists of three modules: 
(a) a deep feature extraction module which extracts object features from the frontal-view visual observations and the ego vehicle state information; 
(b) an importance classification module which takes in the extracted feature embeddings and reasons about relations between the objects in the scene and identifies the important ones; 
and (c) an auxiliary ego behavior prediction module which enhances important object identification by providing additional supervision.
We employ a modified semi-supervised learning algorithm to enable the model to learn from a combination of labeled and unlabeled dataset, which achieves better performance compared with standard supervised learning.

\vspace{-0.cm}
\subsection{Deep Feature Extraction}
\vspace{-0.cm}
We assume that the bounding boxes of objects (i.e., traffic participants, traffic lights, stop signs) in the driving videos can be obtained by a detection and tracking system in advance.
The depth images can be obtained by projecting the point cloud to the frontal camera view.
The segmentation maps are obtained by applying DeepLabv3 \cite{chen2017rethinking} to the RGB images.
The depth images and segmentation maps have the same size as the original RGB images, which are concatenated along the channel axis.
The state information (i.e., position, velocity, acceleration) of the ego vehicle can be obtained from the synchronized 3D LiDAR SLAM and CAN bus. The ego intention $I_E$ is a prior knowledge based on ego planning.

\textbf{Visual Features}:
Considering the object $j$, the visual features $\mathbf{v}_{j, V}$ of a certain object consist of appearance features $\mathbf{v}_{j, A}$ extracted from RGB images, and depth/semantic features $\mathbf{v}_{j, DS}$ extracted from depth images and semantic segmentation maps.
The appearance features contain the information of both the appearance and the local context of objects.
We adopt the ResNet101 \cite{he2016deep} pre-trained on the ImageNet dataset with Feature Pyramid Networks \cite{Lin_2017_CVPR} on top as the backbone of Visual Feature Extractor I to obtain the features at all frames, which are fed into Sequence Encoder I to obtain the final features $\mathbf{v}_{j, A}$.
We use Visual Feature Extractor II (i.e., ResNet18) trained from scratch to extract the depth/semantic features at all frames, which are fed into Sequence Encoder III to obtain the final features $\mathbf{v}_{j, DS}$.
To extract the feature of each object from $\mathbf{v}_{j, A}$ and $\mathbf{v}_{j, DS}$, a ROIAlign \cite{he2018mask} pooling layer is added before feeding into sequence encoder.
The final visual feature of each object is concatenated by $\mathbf{v}_{j, V} = [\mathbf{v}_{j, A}, \mathbf{v}_{j, DS}]$ along the channel dimension.
Similar procedures are applied to extract the global context information from the whole image. 
The global feature is denoted as $\mathbf{v}_{\text{global}} = [\mathbf{v}_{\text{global},A}, \mathbf{v}_{\text{global},DS}]$.

\textbf{Bounding Box Features}:
The location and scale of the object bounding boxes in the frontal-view images can provide additional indications of the size and relative positions of the objects with respect to the ego vehicle, which may influence their importance.
We represent this information by $(\frac{x_{j,t}}{W}, \frac{y_{j,t}}{H}, \frac{w_{j,t}}{W}, \frac{h_{j,t}}{H})$ where $x_{j,t}$, $y_{j,t}$, $w_{j,t}$ and $h_{j,t}$ denote the center coordinate, width and height of the bounding box, $W$ and $H$ denote the width and height of the original image.
The stack of this vector along the time axes is fed into Sequence Encoder II to obtain the bounding box feature $\mathbf{v}_{j,B}$.

\textbf{Ego Vehicle Features}:
We extract the ego state features $\mathbf{v}_{\text{ego}}$ from a sequence of state information (i.e., position, velocity, acceleration) with the Ego State Feature Encoder.

\vspace{-0.cm}
\subsection{Importance Classification on Relation Graph}
\vspace{-0.cm}

After obtaining the ego state features ($\mathbf{v}_{\text{ego}}$), global features ($\mathbf{v}_{\text{global}}$) and object features ($\mathbf{v}_j, j=1,...,N_i$) in the image sequence $\mathbf{I}_i$, we can construct a fully-connected object relation graph where the node attributes are the corresponding object features. 
In order to model the mutual influence and relations among individuals, we apply a message passing mechanism over the graph, which consists of an edge update ($v \to e$) and a node update ($e \to v$):
\begin{align}
	 \mathbf{e}_{jk} = f_e\left([\mathbf{v}_j, \mathbf{v}_k]\right), \quad
	 \bar{\mathbf{v}}_j = f_v\left(\sum\nolimits_{j \neq k} \mathbf{e}_{jk}\right),
\end{align}
where $\mathbf{e}_{jk}$ is the edge attribute from the sender node $k$ to the receiver node $j$, $\bar{\mathbf{v}}_j$ is defined as the relation features of node $j$, and $f_e(\cdot)$/$f_v(\cdot)$ are the edge/node update functions (i.e., multi-layer perceptrons) which are shared across the whole graph.
Note that the edges between the same pair of nodes with opposite directions have different attributes since their mutual influence of object importance may not be symmetric.
The message passing procedure is applied multiple times to model higher-order relations in our model. 

Since the importance of a certain object with respect to the ego vehicle at the current frame not only depends on its own state but also the global context, object relations and the ego vehicle intention, we generate a comprehensive object feature $\mathbf{o}_j = \left[\mathbf{v}_j, \bar{\mathbf{v}}_j, \mathbf{v}_{\text{global}}, \mathbf{v}_{\text{ego}}, I_E \right]$,
which is fed into the classifier (i.e., multi-layer perceptron) to obtain its importance score $s_j \in [0,1]$ (i.e., the probability that the object is important).
During training phase, $s_j$ is used to compute loss directly for labeled objects and generate pseudo-labels for unlabeled ones.
During testing phase, $s_j$ is used to predict importance by $\arg\max (1-s_j,s_j)$.

\subsection{Ranking-Based Pseudo-Label Generation}
\vspace{-0.cm}

Pseudo-label generation is generally a crucial step in semi-supervised learning algorithms.
In our task, a naive way is to use the learned importance classifier at the last iteration directly to assign pseudo-labels for the objects in the unlabeled data samples by $\arg \max (1-s_j, s_j)$. 
However, in most cases only a small subset of objects are important, the naive version of pseudo-label assignment may lead to an data imbalance issue (i.e., assigning an "unimportant" label to all the objects) which degrades the model performance.
In order to mitigate this problem, we adopt a modified ranking-based strategy similar to \cite{hong2020learning}, which encourages the model to identify relative importance.
The advantage of our strategy over \cite{hong2020learning} is to better leverage the confident scores predicted by the model and consider the ranking scores with in two stages with separate thresholds.

First, we label the objects with a raw importance score $s_j$ larger than a threshold $\alpha_1=0.8$ as important objects. Similarly, those with a raw importance score $s_j$ smaller than a threshold $1- \alpha_1$ are labeled as unimportant objects.
If all the objects in a certain case are labeled, there is no further operation and the data sample is appended to the training set.
Second, in order to consider the relative importance of the rest objects, we calculate a set of ranking scores $\bar{s}_j$ via dividing the raw scores $s_j$ by their maximum.
Then we label the objects with a ranking score $\bar{s}_j$ greater than a threshold $\alpha_2=0.8$ as important ones while the others as unimportant ones. This ranking strategy can not only incorporate relational reasoning in the pseudo-label generation process, but also mitigate data imbalance issue to some extent.
We denote the pseudo-label of object $\mathbf{x}^\mathcal{U}_{j}$ as $\tilde{y}^{\mathcal{U}}_j$.

\vspace{-0.cm}
\subsection{Auxiliary Tasks: Ego Vehicle Behavior Prediction}
\vspace{-0.cm}

Since the behavior of ego vehicle can be affected by the existence of important objects, it could in turn enhance the identification of important objects. 
Therefore, we propose to predict the ego vehicle behavior at two levels as parallel auxiliary tasks to provide additional supervision signals. 
Since the ego information is always available without the requirement of human annotations, the auxiliary tasks are trained by supervised learning with ground truth for both labeled and unlabeled datasets.
First, we predict the high-level action of the ego vehicle $\mathcal{A}_E$ at the current frame with the Ego Action Classifier (EAC), which is formulated as a classification problem. 
The ego actions can be \textit{stop}, \textit{speed up}, \textit{slow down} or \textit{constant speed}. 
The ground truth actions are automatically obtained by setting thresholds on the speed and acceleration of the ego vehicle.
Second, we also forecast the low-level trajectory of the ego vehicle $\mathcal{T}_E$ in the future 2 seconds with the Ego Trajectory Generator (ETG), which is formulated as a regression problem.

The EAC and ETG share the same input, which is a feature embedding including the ego intention, ego state features, important object features and global features while discarding the information of unimportant objects.
The intuition is that the ego behavior is only influenced by the important objects, which serves as a regularization to encourage the model to identify the correct ones. 
If some important objects are mistakenly discarded, the predicted ego behavior may change accordingly which results in a discrepancy from the true behavior.
The corresponding loss can help improve the importance classifier.
However, since hard assignment is not differentiable, we employ the Gumbel-Softmax technique \cite{jang2016categorical} to obtain the gradient approximation for back-propagation. 
More formally, we denote the weight associated to object $\mathbf{x}_j$ as $z_j$ which can be drawn as
\begin{equation}
    z_j = \frac{e^{((\log (s_j) + g_{j,1})/\tau)}}{e^{ ((\log (s_j) + g_{j,1})/\tau)} + e^{ ((\log (1-s_j) + g_{j,0})/\tau)}},
\end{equation}
where $\mathbf{g}_j \in \mathbb{R}^2$ is a vector of i.i.d. samples drawn from a $\text{Gumbel}(0,1)$ distribution and $\tau$ is the softmax temperature which controls the smoothness of samples. 
This distribution converges to one-hot samples from the categorical distribution as $\tau \to 0$.
Then the important object features $\mathbf{v}_{\text{imp}}$ is
\begin{align}
	&\mathbf{v}_{\text{imp}} = \frac{1}{N_i} \sum_{j=1}^{N_i} z_j \mathbf{v}_j \ (\text{training}) \\
	\mathbf{v}_{\text{imp}} = \frac{1}{\hat{N}_i} & \sum_{j=1}^{\hat{N}_i} \arg \max (1-s_j, s_j) \mathbf{v}_j \ (\text{testing}),
\end{align}
where $\hat{N}_i$ is the total number of predicted important objects.
The combined feature for ego behavior prediction $[\mathbf{v}_{\text{imp}}, \mathbf{v}_{\text{ego}}, \mathbf{v}_{\text{global}}, I_E]$ is used to predict the ego action $\mathcal{A}_E$ and trajectory $\mathcal{T}_E$.

\vspace{-0.cm}
\subsection{Loss Function and Training}
\vspace{-0.cm}

The proposed model can be trained either by supervised learning on the labeled dataset $\mathcal{L}$ or by semi-supervised learning on the combined (labeled and unlabeled) dataset $\mathcal{L}+\mathcal{U}$. The former one serves as an ablative baseline.

\textbf{Supervised Learning}:
The loss function $L_{\text{SL}}$ consists of two parts: importance classification loss $L_{\text{imp}}$ and auxiliary loss $L_{\text{aux}}$.
More specifically, the loss is computed by
\begin{align}
	L^{\mathcal{L}}_{\text{SL}} =& \ L^{\mathcal{L}}_{\text{imp}} + \lambda L^{\mathcal{L}}_{\text{aux}}\nonumber 
	= - \frac{1}{|\mathcal{L}|}\sum_{i=1}^{|\mathcal{L}|} \frac{1}{N^{\mathcal{L}}_i} \sum_{j=1}^{N^{\mathcal{L}}_i} l_{\text{CE}}(y^\mathcal{L}_j, s^\mathcal{L}_j) + \\
	&\lambda \left(-\frac{1}{|\mathcal{L}|} \sum_{i=1}^{|\mathcal{L}|} l_{\text{CE}}({\mathcal{A}_E}_i, {{}\hat{\mathcal{A}}_{E}}_i)
	+ \beta ||{\mathcal{T}_E}_i - {{}\hat{\mathcal{T}}_E}_i||^2 \right), 
\end{align}
where $l_{\text{CE}}(\cdot,\cdot)$ denotes the binary cross-entropy (CE) loss. $\lambda$ and $\beta$ are used to adjust the ratio between different losses.

\textbf{Semi-Supervised Learning}:
The loss function $L_{\text{SSL}}$ consists of two parts: labeled data loss $L^{\mathcal{L}}$ and unlabeled data loss $L^{\mathcal{U}}$. More specifically, the loss is computed by
\begin{align}
	L_{\text{SSL}} = \ L^{\mathcal{L}}_{\text{SL}} + \gamma L^{\mathcal{U}}_{\text{SL}}, \
	w_j = \frac{\exp s^{\mathcal{U}}_j}{\sum_{j=1}^{N^{\mathcal{U}}_i} \exp s^{\mathcal{U}}_j}, 
	\varepsilon_i = 1 - \frac{H(\mathbf{w})}{H(\mathbf{m})}, 
\end{align}
\vspace{-0.5cm}
\begin{align}
	L^{\mathcal{U}}_{\text{SL}} 
	=& \frac{1}{|\mathcal{U}|} \sum_{i=1}^{|\mathcal{U}|} \varepsilon_i \sum_{j=1}^{N^{\mathcal{U}}_i} w_j l_{\text{MSE}}(\tilde{y}^\mathcal{U}_j, s^\mathcal{U}_j) \nonumber \\
	+& \lambda \left(-\frac{1}{|\mathcal{U}|} \sum_{i=1}^{|\mathcal{U}|}  l_{\text{CE}}({\mathcal{A}_E}_i, {{}\hat{\mathcal{A}}_{E}}_i)
	+ \beta ||{\mathcal{T}_E}_i - {{}\hat{\mathcal{T}}_E}_i||^2 \right),
\end{align}
where $l_{\text{MSE}}(\cdot,\cdot)$ denotes the mean square error (MSE) loss \cite{berthelot2019mixmatch,oliver2018realistic}, $\gamma$ is used to balance the ratio between labeled and unlabeled data loss, $\mathbf{w} = (w_1,...,w_{N^{\mathcal{U}}_i})$, $\mathbf{m} = (1/N^{\mathcal{U}}_i,...,1/N^{\mathcal{U}}_i)$ with the same dimension as $\mathbf{w}$, and $H(\cdot)$ is the entropy function.
$\varepsilon_i$ and $w_j$ are the weights associated to $i$-th unlabeled data case and the $j$-th objects in a certain case, respectively. 
The weight $\gamma$ is initialized as 0, which implies that the unlabeled dataset is not used at the beginning of training. 
It increases to a maximum value over a fixed number of epochs with a linear schedule \cite{berthelot2019mixmatch,oliver2018realistic} since the model becomes more accurate and confident thus can generate more reliable pseudo-labels as training goes on. 

\textbf{Object Weighting}: We apply a weighting mechanism on the loss of each object based on the corresponding predicted importance score $s^{\mathcal{U}}_j$ to strengthen the effect of important ones while weaken that of unimportant ones.

\textbf{Unlabeled Data Weighting}: The basic assumption of our task is that a small subset of objects in the scene are significantly more important to the ego vehicle than the others in most scenarios. In some situations, however, the model may predict vague and similar importance scores for all the objects in unlabeled scenarios. Such cases contribute little to important object identification thus their effects should be weakened. More specifically, we obtain the weight $\varepsilon_i$ by leveraging the entropy function $H(\cdot)$ to indicate the similarity between importance scores of different objects. A set of more similar scores will lead to a smaller weight associated with the corresponding training case.

\section{Experiments}
\vspace{-0.cm}

\subsection{Dataset and Preprocessing}
\vspace{-0.cm}

We experimented with a public driving dataset to validate our model: Honda 3D dataset (H3D) \cite{patil2019h3d}.
The dataset provides the information collected by a full sensor suite (e.g., camera, LiDAR, radar) mounted on a testing vehicle that navigates in complex driving scenarios with highly interactive traffic participants and the annotated bounding boxes of detected objects in the frontal-view images.
Note that in this work we directly used the ground truth bounding boxes provided in the dataset instead of including an upstream detection module. 
The purpose is to focus on the important object identification while minimizing the influencing factors from the upstream perception task.

\textbf{Dataset Statistics}:
We have 7,517 labeled cases for supervised learning and 4,241 unlabeled cases for semi-supervised learning.
The ratio of going forward/turning left/turning right is around 4:1:1.
The labeled cases are randomly split into training/testing datasets with a ratio of 7:3. 

\textbf{Object Importance Annotation}:
In order to obtain the binary importance labels of the object in each bounding box, a group of annotators (i.e., experienced drivers) were asked to watch the egocentric driving videos and imagine themselves as the driver of ego vehicle.
A segment of ego future trajectory was also provided to the annotators for intention annotation. 
For each video segment, the important objects and ego intention were annotated at a frequency of 2Hz.
We define the important objects as the ones that can potentially influence the ego vehicle behavior given a certain ego intention/goal, which includes dynamic/static traffic participants, traffic lights and stop signs.
Generally, the traffic lights and stop signs related to the ego vehicle are labeled as important objects since they provide crucial semantic information of the environment. 
The static/dynamic traffic participants that are located at the potential future path of the ego vehicles are important.
The parked vehicles close to the ego vehicles are generally labeled as important objects since they may start to move and merge into the ego lane.

To demonstrate the consistency and validity of the annotations, we compute the intra-class correlation coefficient (ICC) \cite{shrout1979intraclass}, which is widely used for the assessment of consistency or reproducibility of quantitative measurements made by different observers measuring the same quantity.
The ICC for ego intention annotations is 0.941, and the ICC for important object annotations is 0.934. 
According to the guideline in \cite{cicchetti1994guidelines}, an ICC score between 0.75 and 1.00 indicates ``excellent'', which implies the consistency of our annotations.

\vspace{-0.cm}
\subsection{Evaluation Metrics and Baselines}
\vspace{-0.cm}

We adopt the standard metrics for binary classification problems (i.e., accuracy, F1 score) to evaluate the performance of important object identification.
To the best of our knowledge, our work is the first to consider the influence of traffic lights and stop signs for this specific task, thus no existing baseline is available to compare against directly.
Instead, we conduct a detailed ablative study to demonstrate the effectiveness of each model component, and compare our approach with some rule/heuristics based methods.

The different model settings are elaborated in Table \ref{tab:baselines}.
Ours-S denotes our model trained only on labeled data by standard supervised learning. 
Ours-S-X denotes a variant of Our-S without certain model components or input information.
Ours-SS denotes our model trained both on labeled and unlabeled data by semi-supervised learning.
Ours-SS-X denotes a variant of Our-SS without certain model components.
The check marks (\checkmark) indicate containing the corresponding information or component.
The dashes (--) indicate "not applicable".
We also implement four rule-based baseline methods.
B-1 selects the object with the largest bounding box.
B-2 selects the object closest to the image center.
B-3 selects the object closest to ego vehicle. 
B-4 is the learning-based method proposed in \cite{zhang2020interaction}, which is trained on our dataset.

\setlength{\tabcolsep}{6pt} 
\begin{table}[!tbp]
	\centering
	\caption{The Ablative Baseline Model Settings}
	\vspace{-0.2cm}
	\fontsize{10}{8}\selectfont
	\resizebox{\columnwidth}{!}{
		\begin{tabular}{m{1.5cm}<{\centering}|m{1.7cm}<{\centering}| m{1.7cm}<{\centering}| m{1.7cm}<{\centering}|| m{1.7cm}<{\centering}| m{1.7cm}<{\centering}}
			\toprule
			\midrule
			Model  & Ego Intention & Relation Graph & Auxiliary Task & Ranking Strategy & Pseudo Loss Weighting  \\
			\midrule
			Ours-S-1 & \checkmark &  &  & -- & --  \\
			Ours-S-2 &  & \checkmark &  & -- & --  \\
			Ours-S-3 & \checkmark & \checkmark &  & -- & --  \\
			Ours-S & \checkmark & \checkmark & \checkmark & -- & --  \\ 
			\midrule
			Ours-SS-1 & \checkmark & \checkmark &  & \checkmark &   \\
			Ours-SS-2 & \checkmark & \checkmark &  &  & \checkmark  \\
			Ours-SS-3 & \checkmark & \checkmark &  & \checkmark & \checkmark  \\
			Ours-SS & \checkmark & \checkmark & \checkmark & \checkmark & \checkmark  \\
			\bottomrule
		\end{tabular}
	}
	\label{tab:baselines}
\vspace{-0.7cm}
\end{table}

\setlength{\tabcolsep}{2pt} 
\begin{table*}[!tbp]
	\caption{Comparison of Important Object Identification Performance (\%) on H3D Dataset}
	\vspace{-0.5cm}
	\fontsize{12}{10}\selectfont
	\begin{center}
		\resizebox{0.99\textwidth}{!}{
			\begin{tabular}{m{2.2cm}<{\centering}|m{1.8cm}<{\centering}||m{1.8cm}<{\centering}|m{1.8cm}<{\centering}|m{1.8cm}<{\centering}|m{1.8cm}<{\centering}||m{1.8cm}<{\centering}|m{1.8cm}<{\centering}| m{1.8cm}<{\centering}| m{1.8cm}<{\centering}| m{1.8cm}<{\centering}|| m{1.8cm}<{\centering}| m{1.8cm}<{\centering} | m{1.8cm}<{\centering}| m{1.8cm}<{\centering}}
				\toprule
				\midrule
				Ego Intention & Metric  & B-1 & B-2 & B-3 & B-4 \cite{zhang2020interaction} & Ours-S w/o depth/seg &Ours-S-1 & Ours-S-2 & Ours-S-3 & Ours-S & Ours-SS-1 & Ours-SS-2 & Ours-SS-3 & Ours-SS \\
				\midrule 
				\multirow{2}*{\makecell[c]{Left  Turn \\(LT)}}  
				&Accuracy	 & 78.1 & 67.3 & 76.1& 76.8 & 71.7 &79.9 & 76.3 & 81.0 & 82.4 & 81.3 & 81.3 & 81.7 & \textbf{87.6} \\ 
				&F1 Score	 & 43.4 & 15.6 & 38.2 & 65.6 & 62.6 & 64.1 & 63.7 & 66.7 & 68.2 & 68.4 & 68.2 & 68.2 & \textbf{73.2} \\
				\midrule
				\multirow{2}*{\makecell[c]{Straight  Pass \\(SP)}}
				&Accuracy	 & 79.1 & 76.9 & 75.0 & 78.1 & 68.5 & 79.0 & 75.0 & 82.2 & 83.6 & 84.3 & 84.7 & 85.1 & \textbf{89.5}\\ 
				&F1 Score	 & 33.8 & 27.0 & 21.0 & 66.0 & 50.5 & 59.2 & 56.0 & 65.8 & 68.7 & 67.7 & 69.2 & \textbf{73.2} & 71.8 \\
				\midrule
				\multirow{2}*{\makecell[c]{Right  Turn \\(RT)}}   
				&Accuracy	 & 70.9 & 62.1 & 72.6 & 74.9 & 57.1 & 74.7 & 72.7 & 75.8 & 78.0 & 78.6 & 78.7 & 80.7 & \textbf{82.6} \\  
				&F1 Score	 & 33.8 & 23.1 & 44.4 & 67.3 & 58.6 & 62.4 & 59.1 & 68.6 & 70.2 & 72.3 & 69.2& \textbf{74.0} & 72.3\\
				\midrule
				\multirow{2}*{\makecell[c]{All}}   
				&Accuracy	 & 77.6 & 73.2 & 74.8 & 77.4 & 67.1 & 78.5 & 75.5 & 81.0 & 82.6 & 82.9 & 83.3 & 83.9 & \textbf{88.1} \\ 
				&F1 Score    & 36.8 & 24.4 & 28.8 & 66.5 & 54.1 & 60.5 & 56.4 & 66.6 & 68.9 & 68.9 & 69.1 & 70.5 & \textbf{72.1} \\
				\bottomrule
			\end{tabular}
		}
	\end{center}
	\vspace{-0.8cm}
	\label{tab:H3D}
\end{table*}

\setlength{\tabcolsep}{2pt} 
\begin{table}[!tbp]
	\caption{Semi-Supervised Learning Performance (\%) on H3D Dataset}
	\vspace{-0.5cm}
	\fontsize{10}{8}\selectfont
	\begin{center}
		\resizebox{0.9\columnwidth}{!}{
			\begin{tabular}{m{1.8cm}<{\centering}|m{1.8cm}<{\centering}||m{2cm}<{\centering}| m{2cm}<{\centering}|m{2cm}<{\centering}}
				\toprule
				\midrule
				Ego Intention & Metric  & Ours-SS-MT \cite{tarvainen2017mean} & Ours-SS-NS \cite{xie2020self} &  Ours-SS \\
				\midrule 
				\multirow{2}*{\makecell[c]{All}}   
				&Accuracy	  & 87.9 & 80.42 & \textbf{88.1}\\ 
				&F1 Score	  & 68.9 & 65.11 & \textbf{72.1}\\
				\bottomrule
			\end{tabular}
		}
	\end{center}
	\vspace{-0.5cm}
	\label{tab:SSL}
\end{table}

\subsection{Implementation Details}
For all the experiments, a batch size of 32 was used and the models were trained for up to 100 epochs with early stopping.
We used Adam optimizer with an initial learning rate of 0.0001. 
The models were trained on a single NVIDIA Quadro V100 GPU.
We set the thresholds for pseudo-label generation as $\alpha_1 = 0.8 ,\alpha_2 = 0.8$ and the softmax temperature as $\tau = 0.1$.
In the loss functions, we set $\lambda = 0.5$, $\beta = 1.0$, $\gamma$ is initialized as 0.001 and exponentially increases at each iteration to its maximum value 1.0.
Specific details of model components are introduced below:
(a) Sequence Encoder I/II/III and Ego State Feature Encoder: a single-layer LSTM with hidden size = 128 followed by a three-layer MLP with hidden size = 128. The dimensions of all types of output features are 128.
(b) Edge/Node Update Functions $f_v(\cdot)$/$f_e(\cdot)$: a three-layer MLP with hidden size = 128.
(c) Importance Classifier: a three-layer MLP with hidden size = 256.
(d) Ego Action Classifier and Ego Trajectory Generator: a three-layer MLP with hidden size = 64 followed by a three-layer MLP with hidden size = 256.

\vspace{-0.cm}
\subsection{Results and Analysis}
\vspace{-0.cm}

\begin{figure}[!tbp]
	\centering
	\includegraphics[width=0.95\columnwidth]{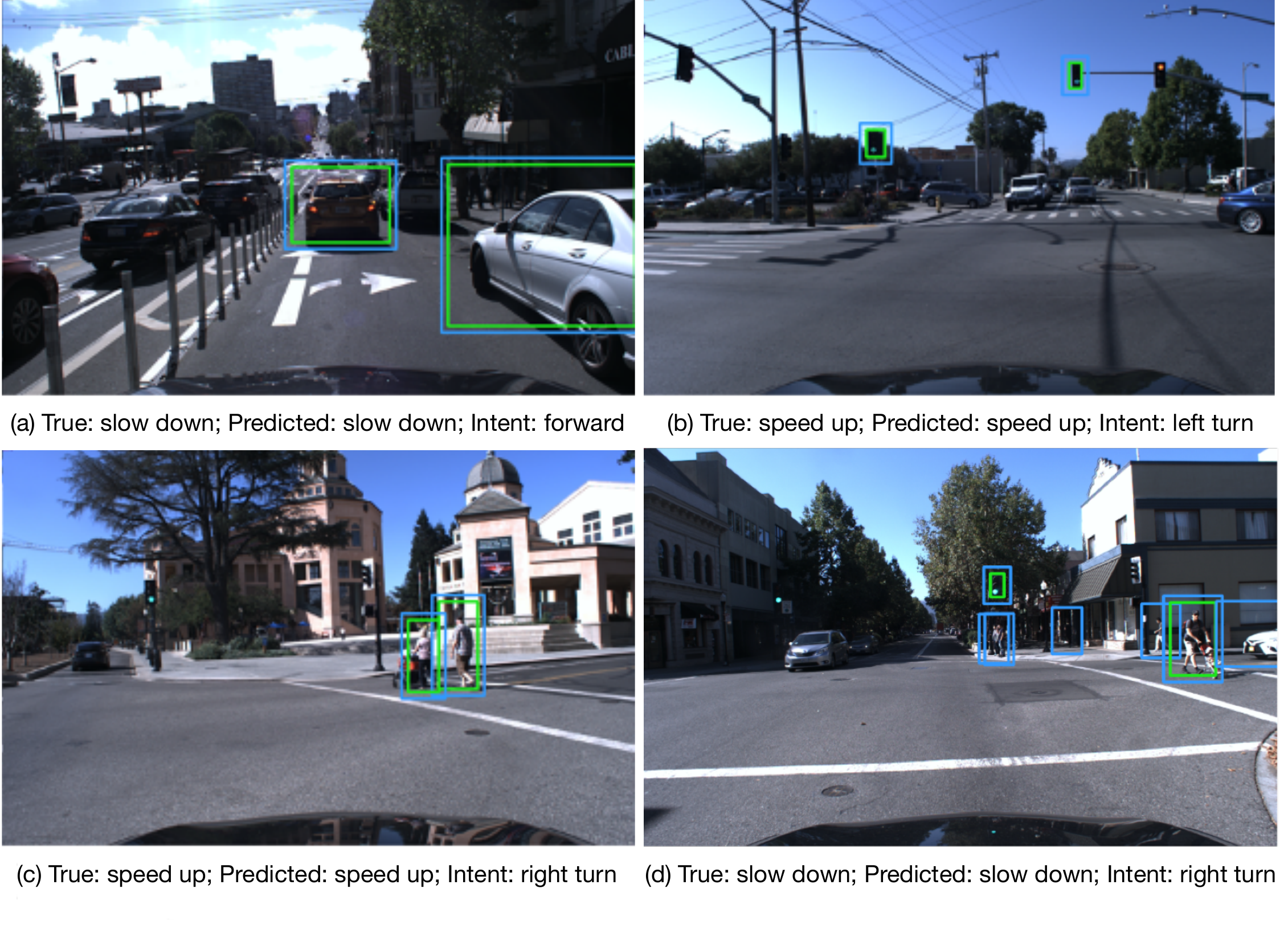}
	\vspace{-0.6cm}
	\caption{Testing scenarios with true ego vehicle intentions (Ours-SS). The green/blue boxes indicate the true/predicted important objects. Our model can identify the correct important objects and predict accurate ego actions. In (a), the model recognizes adjacent objects as important due to potential interactions. 
	In (b), since the light for ``turning left'' is green, the vehicles in the opposite direction are not important thus the ego vehicle can speed up.
	In (c), the model identifies the important objects in the intended path of ego vehicle. 
	In (d), since the ego intention is ``turning right'', the model predicts the objects in the right part of the image as important ones and still predicts the ego action correctly and keeps the ego vehicle safe.}
	\vspace{-0.6cm}
	\label{fig:plot_test1}
\end{figure}

\begin{figure}[!tbp]
	\centering
	\vspace{-0.cm}
	\includegraphics[width=0.75\columnwidth,height=6.5cm]{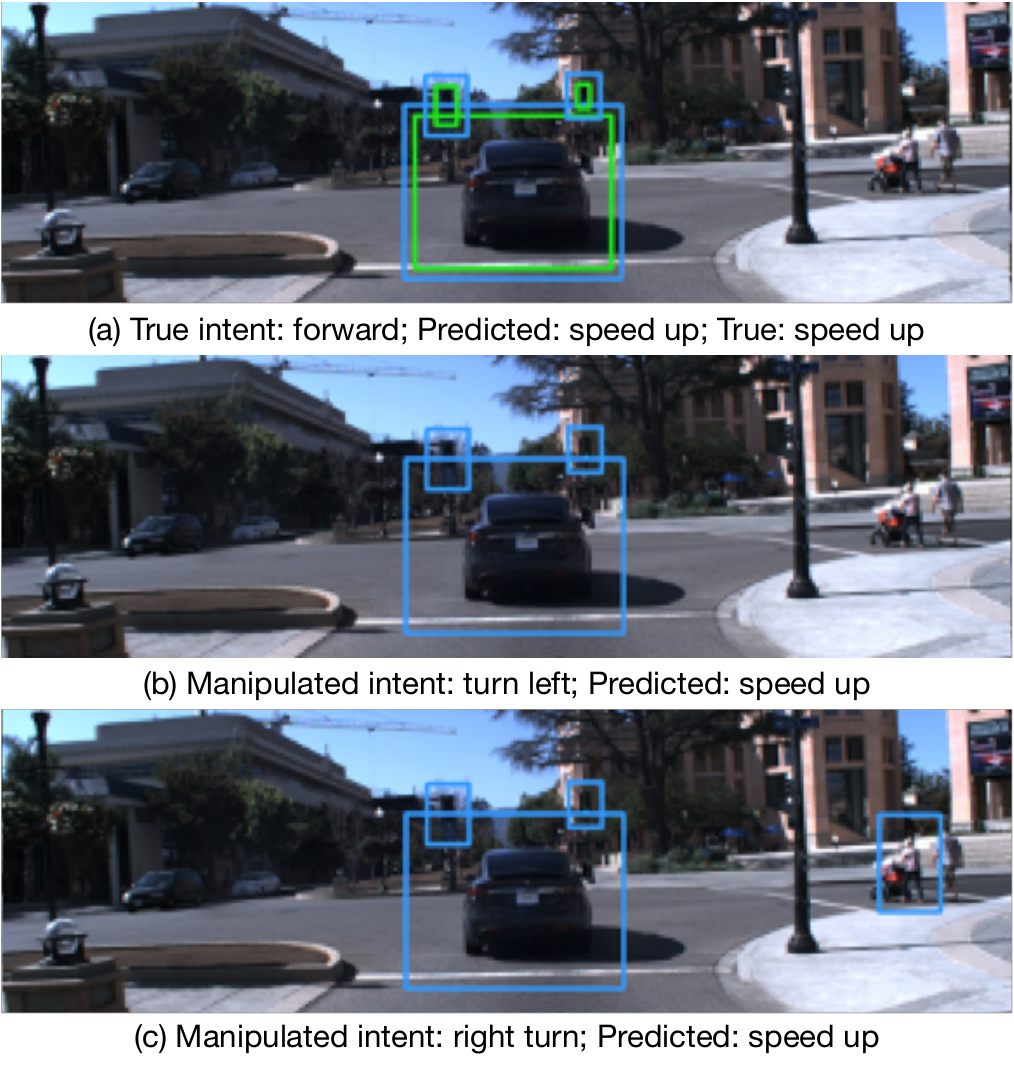}
	\vspace{-0.4cm}
	\caption{Testing scenarios with manipulated ego vehicle intentions (Ours-SS). Blue boxes denote predicted important objects and green ones denote the ground truth. The model makes reasonable predictions with both true and manipulated ego intents.}
	\vspace{-0.4cm}
	\label{fig:plot_test2}
\end{figure}

We conducted a detailed ablation study with quantitative and qualitative results to demonstrate the effectiveness of each model component in the supervised learning setting and effectiveness of the ranking-based pseudo-label generation in the semi-supervised learning setting.
We also compared our method with some rule/heuristics-based baselines.
The numerical comparisons of performance are provided in Table \ref{tab:H3D}.
Some testing scenarios are visualized in Fig. \ref{fig:plot_test1} and Fig. \ref{fig:plot_test2}.

\textbf{Rule/Learning-based baselines}:
We demonstrate the advantages of our method over rule-based methods and the learning-based method proposed in \cite{zhang2020interaction} in Table \ref{tab:H3D}. We can observe a large drop in performance by using various heuristics to identify the important objects, which implies that fixed rules are not sufficient to capture complex relations and relative importance of objects in the scene.

\textbf{Ego Intention}:
In a certain scenario, the important objects may be distinct with different ego driving intentions. 
We evaluated the model performance on the scenarios with different ego intentions separately to illustrate its effectiveness on H3D dataset.
Comparing Ours-S-2 and Ours-S-3 in both tables, the accuracy and F1 score improve with ego intention by a large margin consistently in the scenarios with various ego intentions. 
More specifically, leveraging the ego intention improves accuracy by 5.5\% and F1 score by 10.2\% in average.
The intention information can guide the model to pay attention to specific regions thus helpful for important object identification.
In order to qualitatively illustrate the influence of ego intention, we visualize the predicted important objects with both ground truth and manipulated intentions in Fig. \ref{fig:plot_test2}. The results show that our model can not only correctly identify the important objects with the true intention, but also generate reasonable predictions for manipulated intentions with which the attention may be paid to the objects in different corresponding regions.

\textbf{Relation Graph}:
The relation graph is a critical component to aggregate the information of different objects and model their relations/interactions, which helps to discriminate relative importance of multiple objects.
In Ours-S-1, the relation graph is skipped and the relation features $\bar{\mathbf{v}}_j$ is removed from $\mathbf{o}_j$, which leads to a decrease in performance compared with Ours-S-3, which implies the effectiveness of graph learning. The information aggregation of different objects boosts the model accuracy by 2.5\% and F1 score by 6.1\%.

\textbf{Auxiliary Tasks}:
The comparison between Ours-S and Ours-S-3, Ours-SS and Ours-SS-3 can illustrate the effectiveness of the auxiliary tasks.
The results show that adding auxiliary tasks during training improves accuracy and F1 score in both supervised learning and semi-supervised learning, which implies that the ego behavior indeed provides useful information and additional supervision.
In Fig. \ref{fig:plot_test1}, it shows that our model can predict accurate ego actions at the current frame in various scenarios with specific intentions.

\textbf{Semi-Supervised Learning}:
The comparison between Ours-S and Ours-SS shows that the performance of our model improves by a large margin with semi-supervised learning. The accuracy/F1 score improve by 5.5\%/3.2\%, respectively.
The ablative results in Table \ref{tab:H3D} also illustrate the benefits brought by ranking-based strategy for pseudo-label generation and entropy-based loss weighting.
Our method outperforms baseline semi-supervised learning methods in Table \ref{tab:SSL}.

\begin{figure}[!tbp]
\centering
{\includegraphics[width=1.02\columnwidth,height=3cm]{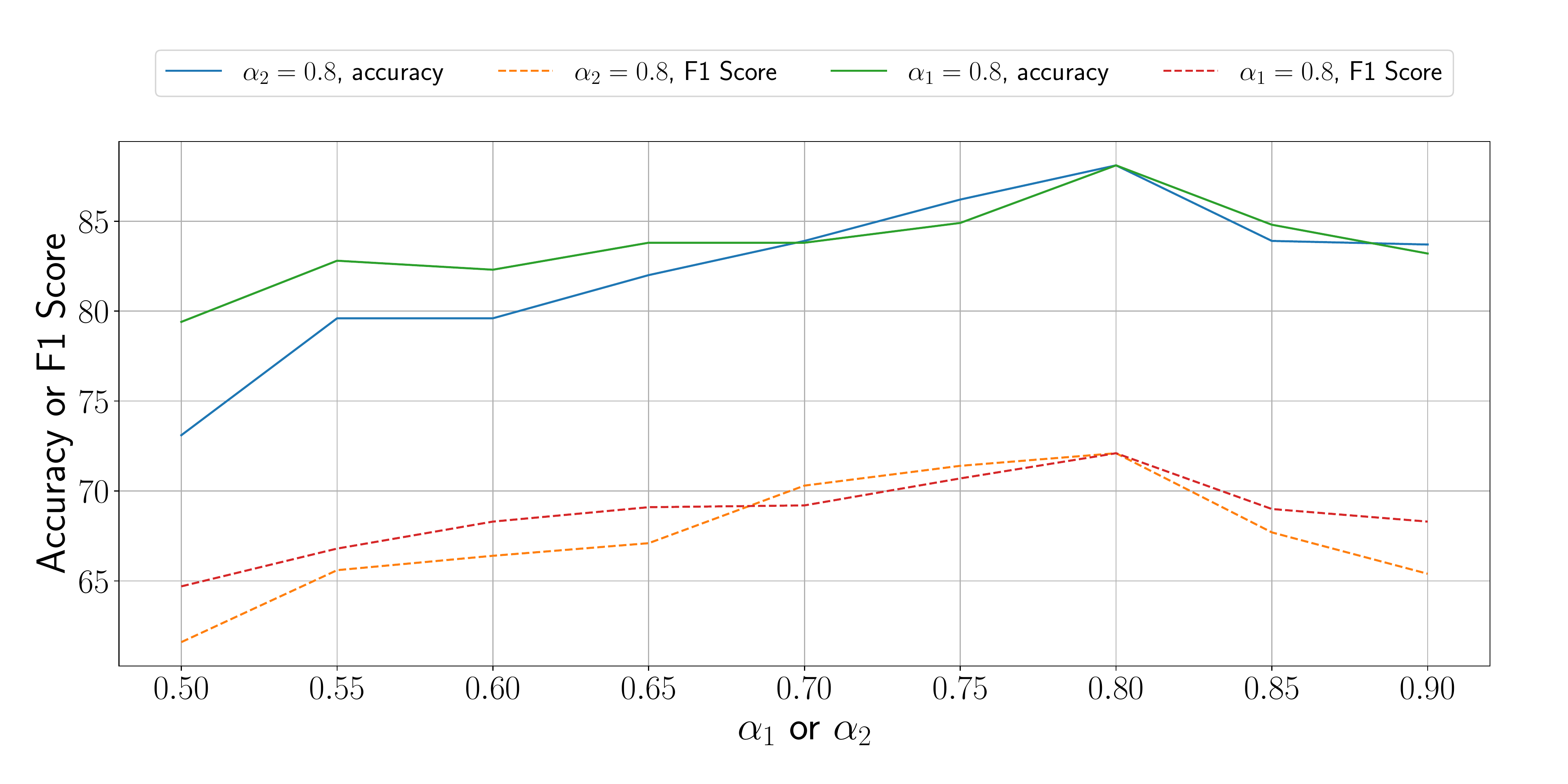}}
\vspace{-0.8cm}
\caption{The comparison of model performance with different $\alpha_1$ and $\alpha_2$.}
\vspace{-0.6cm}
\label{fig:alpha}
\end{figure}

\textbf{Effect of Thresholds $\alpha_1$ and $\alpha_2$}:
In the pseudo-label generation process, the thresholds $\alpha_1$ and $\alpha_2$ are determined empirically. The comparison of model performance with different $\alpha_1$/$\alpha_2$ combinations is shown in Fig. \ref{fig:alpha}.
In particular, $\alpha_1=0.5$ performs the worst when $\alpha_2$ is fixed at 0.8, which implies that the naive threshold (i.e., 0.5) is not good for generating pseudo labels since the prediction confidence may not be high enough which results in noisy pseudo labels.

\section{CONCLUSIONS}
In this paper, we present a novel approach for important object identification in autonomous driving.
We propose to incorporate human knowledge as direct supervision to recognize important objects.
Since it is expensive to obtain human annotations for a large amount of data, we propose a semi-supervised learning pipeline to enable the model to learn from both labeled and unlabeled datasets.
In order to consider the relations between objects when inferring their importance, we employ a graph neural network to extract relation features.
Moreover, we propose to leverage the action/trajectory information of the ego vehicle to provide additional supervision signals as auxiliary tasks to improve the model performance.
The model is validated on the H3D dataset with egocentric videos, 2D bounding box annotations and point clouds.
The quantitative and qualitative results demonstrate the effectiveness of each model component, the ranking-based pseudo-label generation strategy and loss weighting strategies for semi-supervised learning.

{\small
	\bibliographystyle{IEEEtran}
	\bibliography{egbib}

\begin{thebibliography}{10}
\providecommand{\url}[1]{#1}
\csname url@rmstyle\endcsname
\providecommand{\newblock}{\relax}
\providecommand{\bibinfo}[2]{#2}
\providecommand\BIBentrySTDinterwordspacing{\spaceskip=0pt\relax}
\providecommand\BIBentryALTinterwordstretchfactor{4}
\providecommand\BIBentryALTinterwordspacing{\spaceskip=\fontdimen2\font plus
\BIBentryALTinterwordstretchfactor\fontdimen3\font minus
  \fontdimen4\font\relax}
\providecommand\BIBforeignlanguage[2]{{%
\expandafter\ifx\csname l@#1\endcsname\relax
\typeout{** WARNING: IEEEtran.bst: No hyphenation pattern has been}%
\typeout{** loaded for the language `#1'. Using the pattern for}%
\typeout{** the default language instead.}%
\else
\language=\csname l@#1\endcsname
\fi
#2}}

\bibitem{martin2018dynamics}
S.~Martin, S.~Vora, K.~Yuen, and M.~M. Trivedi, ``Dynamics of driver's gaze:
  Explorations in behavior modeling and maneuver prediction,'' \emph{IEEE
  Trans. on Intell. Vehicles}, vol.~3, no.~2, pp. 141--150, 2018.

\bibitem{palazzi2018predicting}
A.~Palazzi, D.~Abati, F.~Solera, R.~Cucchiara, \emph{et~al.}, ``Predicting the
  driver's focus of attention: the dr (eye) ve project,'' \emph{IEEE Trans. on
  Patt. Analy. and Mach. Intell.}, vol.~41, no.~7, pp. 1720--1733, 2018.

\bibitem{vemula2018social}
A.~Vemula, K.~Muelling, and J.~Oh, ``Social attention: Modeling attention in
  human crowds,'' in \emph{ICRA}.\hskip 1em plus 0.5em minus 0.4em\relax IEEE,
  2018, pp. 4601--4607.

\bibitem{li2021rain}
J.~Li, F.~Yang, H.~Ma, S.~Malla, M.~Tomizuka, and C.~Choi, ``Rain: Reinforced
  hybrid attention inference network for motion forecasting,'' in \emph{ICCV},
  2021, pp. 16\,096--16\,106.

\bibitem{li2021spatio}
J.~Li, H.~Ma, Z.~Zhang, J.~Li, and M.~Tomizuka, ``Spatio-temporal graph
  dual-attention network for multi-agent prediction and tracking,'' \emph{IEEE
  Transactions on Intelligent Transportation Systems}, 2021.

\bibitem{aksoy2020see}
E.~Aksoy, A.~Yaz{\i}c{\i}, and M.~Kasap, ``See, attend and brake: An
  attention-based saliency map prediction model for end-to-end driving,''
  \emph{arXiv preprint arXiv:2002.11020}, 2020.

\bibitem{gao2019goal}
M.~Gao, A.~Tawari, and S.~Martin, ``Goal-oriented object importance estimation
  in on-road driving videos,'' in \emph{ICRA}.\hskip 1em plus 0.5em minus
  0.4em\relax IEEE, 2019.

\bibitem{zhang2020interaction}
Z.~Zhang, A.~Tawari, S.~Martin, and D.~Crandall, ``Interaction graphs for
  object importance estimation in on-road driving videos,'' in
  \emph{ICRA}.\hskip 1em plus 0.5em minus 0.4em\relax IEEE, 2020, pp.
  8920--8927.

\bibitem{anderson2018bottom}
P.~Anderson, X.~He, C.~Buehler, D.~Teney, M.~Johnson, S.~Gould, and L.~Zhang,
  ``Bottom-up and top-down attention for image captioning and visual question
  answering,'' in \emph{CVPR}, 2018, pp. 6077--6086.

\bibitem{yu2019deep}
Z.~Yu, J.~Yu, Y.~Cui, D.~Tao, and Q.~Tian, ``Deep modular co-attention networks
  for visual question answering,'' in \emph{CVPR}, 2019.

\bibitem{sun2020mining}
G.~Sun, W.~Wang, J.~Dai, and L.~Van~Gool, ``Mining cross-image semantics for
  weakly supervised semantic segmentation,'' in \emph{ECCV}.\hskip 1em plus
  0.5em minus 0.4em\relax Springer, 2020, pp. 347--365.

\bibitem{ji2020deep}
Z.~Ji, Y.~Zhao, Y.~Pang, X.~Li, and J.~Han, ``Deep attentive video
  summarization with distribution consistency learning,'' \emph{IEEE Trans. on
  Neural Networks and Learning Systems}, 2020.

\bibitem{li2019learning}
W.-H. Li, F.-T. Hong, and W.-S. Zheng, ``Learning to learn relation for
  important people detection in still images,'' in \emph{CVPR}, 2019.

\bibitem{li2020who}
C.~{Li}, S.~H. {Chan}, and Y.~T. {Chen}, ``Who make drivers stop? towards
  driver-centric risk assessment: Risk object identification via causal
  inference,'' in \emph{IROS}, 2020, pp. 10\,711--10\,718.

\bibitem{zhou2019collaborative}
Y.~Zhou, X.~He, L.~Huang, L.~Liu, F.~Zhu, S.~Cui, and L.~Shao, ``Collaborative
  learning of semi-supervised segmentation and classification for medical
  images,'' in \emph{CVPR}, 2019, pp. 2079--2088.

\bibitem{jing2021videossl}
L.~Jing, T.~Parag, Z.~Wu, Y.~Tian, and H.~Wang, ``Videossl: Semi-supervised
  learning for video classification,'' in \emph{WACV}, 2021.

\bibitem{hong2020learning}
F.-T. Hong, W.-H. Li, and W.-S. Zheng, ``Learning to detect important people in
  unlabelled images for semi-supervised important people detection,'' in
  \emph{CVPR}, 2020, pp. 4146--4154.

\bibitem{ouali2020overview}
Y.~Ouali, C.~Hudelot, and M.~Tami, ``An overview of deep semi-supervised
  learning,'' \emph{arXiv preprint arXiv:2006.05278}, 2020.

\bibitem{tarvainen2017mean}
A.~Tarvainen and H.~Valpola, ``Mean teachers are better role models:
  Weight-averaged consistency targets improve semi-supervised deep learning
  results,'' in \emph{NIPS}, 2017, pp. 1195--1204.

\bibitem{lee2013pseudo}
D.-H. Lee \emph{et~al.}, ``Pseudo-label: The simple and efficient
  semi-supervised learning method for deep neural networks,'' in \emph{Workshop
  on challenges in representation learning, ICML}, vol.~3, no.~2, 2013.

\bibitem{kingma2014semi}
D.~P. Kingma, D.~J. Rezende, S.~Mohamed, and M.~Welling, ``Semi-supervised
  learning with deep generative models,'' in \emph{NIPS}, 2014.

\bibitem{iscen2019label}
A.~Iscen, G.~Tolias, Y.~Avrithis, and O.~Chum, ``Label propagation for deep
  semi-supervised learning,'' in \emph{CVPR}, 2019, pp. 5070--5079.

\bibitem{choi2020shared}
C.~Choi, J.~H. Choi, J.~Li, and S.~Malla, ``Shared cross-modal trajectory
  prediction for autonomous driving,'' in \emph{CVPR}, 2021, pp. 244--253.

\bibitem{li2020evolvegraph}
J.~Li, F.~Yang, M.~Tomizuka, and C.~Choi, ``Evolvegraph: Multi-agent trajectory
  prediction with dynamic relational reasoning,'' \emph{NeurIPS}, vol.~33,
  2020.

\bibitem{ma2021continual}
H.~Ma, Y.~Sun, J.~Li, M.~Tomizuka, and C.~Choi, ``Continual multi-agent
  interaction behavior prediction with conditional generative memory,''
  \emph{IEEE Robotics and Automation Letters}, vol.~6, no.~4, pp. 8410--8417,
  2021.

\bibitem{ma2021multi}
H.~Ma, Y.~Sun, J.~Li, and M.~Tomizuka, ``Multi-agent driving behavior
  prediction across different scenarios with self-supervised domain
  knowledge,'' in \emph{ITSC}.\hskip 1em plus 0.5em minus 0.4em\relax IEEE,
  2021, pp. 3122--3129.

\bibitem{zhou2022grouptron}
R.~Zhou, H.~Gao, H.~Zhou, M.~Tomizuka, J.~Li, and Z.~Xu, ``Grouptron: Dynamic
  multi-scale graph convolutional networks for group-aware dense crowd
  trajectory forecasting,'' in \emph{ICRA}.\hskip 1em plus 0.5em minus
  0.4em\relax IEEE, 2022.

\bibitem{zhou2019relation}
P.~Zhou and M.~Chi, ``Relation parsing neural network for human-object
  interaction detection,'' in \emph{ICCV}, 2019, pp. 843--851.

\bibitem{hu2018relation}
H.~Hu, J.~Gu, Z.~Zhang, J.~Dai, and Y.~Wei, ``Relation networks for object
  detection,'' in \emph{CVPR}, 2018, pp. 3588--3597.

\bibitem{sanchez2020learning}
A.~Sanchez-Gonzalez, J.~Godwin, T.~Pfaff, R.~Ying, J.~Leskovec, and
  P.~Battaglia, ``Learning to simulate complex physics with graph networks,''
  in \emph{ICML}.\hskip 1em plus 0.5em minus 0.4em\relax PMLR, 2020, pp.
  8459--8468.

\bibitem{chen2020relational}
C.~{Chen}, S.~{Hu}, P.~{Nikdel}, G.~{Mori}, and M.~{Savva}, ``Relational graph
  learning for crowd navigation,'' in \emph{IROS}, 2020, pp. 10\,007--10\,013.

\bibitem{battaglia2018relational}
P.~W. Battaglia, J.~B. Hamrick, V.~Bapst, A.~Sanchez-Gonzalez, V.~Zambaldi,
  M.~Malinowski, A.~Tacchetti, D.~Raposo, A.~Santoro, R.~Faulkner,
  \emph{et~al.}, ``Relational inductive biases, deep learning, and graph
  networks,'' \emph{arXiv preprint arXiv:1806.01261}, 2018.

\bibitem{alet2019neural}
F.~Alet, E.~Weng, T.~Lozano-P{\'e}rez, and L.~P. Kaelbling, ``Neural relational
  inference with fast modular meta-learning,'' in \emph{NeurIPS}, 2019, pp.
  11\,804--11\,815.

\bibitem{chen2017rethinking}
L.-C. Chen, G.~Papandreou, F.~Schroff, and H.~Adam, ``Rethinking atrous
  convolution for semantic image segmentation,'' \emph{arXiv preprint
  arXiv:1706.05587}, 2017.

\bibitem{he2016deep}
K.~He, X.~Zhang, S.~Ren, and J.~Sun, ``Deep residual learning for image
  recognition,'' in \emph{ICCV}, 2016, pp. 770--778.

\bibitem{Lin_2017_CVPR}
T.-Y. Lin, P.~Dollar, R.~Girshick, K.~He, B.~Hariharan, and S.~Belongie,
  ``Feature pyramid networks for object detection,'' in \emph{CVPR}, July 2017.

\bibitem{he2018mask}
K.~He, G.~Gkioxari, P.~Doll{\'a}r, and R.~Girshick, ``Mask r-cnn,'' in
  \emph{ICCV}, 2017, pp. 2961--2969.

\bibitem{jang2016categorical}
E.~Jang, S.~Gu, and B.~Poole, ``Categorical reparameterization with
  gumbel-softmax,'' \emph{arXiv preprint arXiv:1611.01144}, 2016.

\bibitem{berthelot2019mixmatch}
D.~Berthelot, N.~Carlini, I.~Goodfellow, N.~Papernot, A.~Oliver, and C.~Raffel,
  ``Mixmatch: A holistic approach to semi-supervised learning,'' \emph{arXiv
  preprint arXiv:1905.02249}, 2019.

\bibitem{oliver2018realistic}
A.~Oliver, A.~Odena, C.~A. Raffel, E.~D. Cubuk, and I.~Goodfellow, ``Realistic
  evaluation of deep semi-supervised learning algorithms,'' \emph{NeurIPS},
  vol.~31, pp. 3235--3246, 2018.

\bibitem{patil2019h3d}
A.~Patil, S.~Malla, H.~Gang, and Y.-T. Chen, ``The h3d dataset for
  full-surround 3d multi-object detection and tracking in crowded urban
  scenes,'' in \emph{ICRA}.\hskip 1em plus 0.5em minus 0.4em\relax IEEE, 2019,
  pp. 9552--9557.

\bibitem{shrout1979intraclass}
P.~E. Shrout and J.~L. Fleiss, ``Intraclass correlations: uses in assessing
  rater reliability.'' \emph{Psychological bulletin}, vol.~86, no.~2, p. 420,
  1979.

\bibitem{cicchetti1994guidelines}
D.~V. Cicchetti, ``Guidelines, criteria, and rules of thumb for evaluating
  normed and standardized assessment instruments in psychology.''
  \emph{Psychological assessment}, vol.~6, no.~4, p. 284, 1994.

\bibitem{xie2020self}
Q.~Xie, M.-T. Luong, E.~Hovy, and Q.~V. Le, ``Self-training with noisy student
  improves imagenet classification,'' in \emph{Proceedings of the IEEE/CVF
  Conference on Computer Vision and Pattern Recognition}, 2020, pp.
  10\,687--10\,698.

\end{thebibliography}
}

\addtolength{\textheight}{-12cm}   


\end{document}